\documentclass[conference]{IEEEtran}
\IEEEoverridecommandlockouts
\usepackage{cite}
\usepackage{amsmath,amssymb,amsfonts}
\usepackage{algorithmic}
\usepackage{graphicx}
\usepackage{textcomp}
\usepackage{xcolor}
\usepackage{booktabs}
\usepackage{float}
\usepackage{multicol}
\usepackage{multirow}
\usepackage{afterpage}

\def\BibTeX{{\rm B\kern-.05em{\sc i\kern-.025em b}\kern-.08em
    T\kern-.1667em\lower.7ex\hbox{E}\kern-.125emX}}
\begin{document}

\title{Harnessing LLMs for API Interactions: A Framework for Classification and Synthetic Data Generation\\
}

\author{\IEEEauthorblockN{Chunliang Tao*}
\IEEEauthorblockA{\textit{New York University}\\
ct1942@nyu.edu}
*Corresponding Author
~\\
\and
\IEEEauthorblockN{Xiaojing Fan}
\IEEEauthorblockA{\textit{New York University}\\
xf435@nyu.edu}

~\\
\and
\IEEEauthorblockN{Yahe Yang}
\IEEEauthorblockA{\textit{George Washington University}\\
yahe.yang@gwu.edu}

}

\maketitle

\begin{abstract}
As Large Language Models (LLMs) advance in natural language processing, there is growing interest in leveraging their capabilities to simplify software interactions. In this paper, we propose a novel system that integrates LLMs for both classifying natural language inputs into corresponding API calls and automating the creation of sample datasets tailored to specific API functions. By classifying natural language commands, our system allows users to invoke complex software functionalities through simple inputs, improving interaction efficiency and lowering the barrier to software utilization. Our dataset generation approach also enables the efficient and systematic evaluation of different LLMs in classifying API calls, offering a practical tool for developers or business owners to assess the suitability of LLMs for customized API management. We conduct experiments on several prominent LLMs using generated sample datasets for various API functions. The results show that GPT-4 achieves a high classification accuracy of 0.996, while LLaMA-3-8B performs much worse at 0.759. These findings highlight the potential of LLMs to transform API management and validate the effectiveness of our system in guiding model testing and selection across diverse applications.
\end{abstract}


\section{Introduction}
In recent years, Large Language Models (LLMs) have made significant advancements in natural language processing (NLP) \cite{vaswani2017attention, devlin2018bert, zhang2022can}, excelling in tasks ranging from text generation to complex problem-solving across industries such as finance, healthcare, arts, and customer services \cite{wu2023bloomberggpt, deng2024composerx, yuan2024rhyme, liu2023text}. These advancements have also sparked a growing exploration into the potential of LLMs to simplify and optimize software interactions. Machine learning and advanced deep learning techniques have been extensively explored to enhance the integration of software systems and optimize a wide range of applications \cite{li2024intelligent, weng2024big, tao2019fact, yukun2019deep}. As a step forward, LLMs are now being studied as powerful tools to make software systems more intuitive and accessible to users of varying technical expertise \cite{ji2024rag, calo2023leveraging}.

Traditionally, users interact with software, particularly through Application Programming Interfaces (APIs), which are crucial for enabling communication between different software applications \cite{fielding2000architectural}. However, interacting with APIs typically requires a solid understanding of their structure, parameters, and specific calls, posing a barrier for non-technical users or those unfamiliar with the API's underlying logic \cite{pautasso2008restful}. Integrating LLMs into API management workflows offers an opportunity to bridge the gap by allowing users to interact with APIs through simple, natural language inputs, which can open up new possibilities for users with varying levels of technical expertise and needs \cite{liang2024taskmatrix, song2023restgpt}.

However, deploying LLMs for API management involves some challenges, primarily in ensuring that the models accurately interpret and classify natural language inputs into the correct API calls. Given that APIs have diverse structures and user inputs vary significantly depending on the context, it is crucial to develop a reliable system for evaluating LLM performance across different use cases. To address the challenges, we propose a novel system that integrates LLMs for two key functionalities. One functionality is to leverage LLMs to interpret and classify natural language inputs and accurately map them to corresponding API calls. Another part is using LLM to automate the generation of sample datasets tailored to specific API functions, which is essential for systematically evaluating LLM performance across various API classification tasks. Unlike conventional approaches, our framework offers a scalable and replicable solution to ensure that API workflows are thoroughly tested with high accuracy and relevance to real-world applications.

We conduct extensive experiments by generating sample datasets using GPT-4-turbo for various API functions and evaluating the classification capabilities of several prominent LLMs, including GPT-4, GPT-4o-mini, LLaMA-3-8B, Gemini-1.5, etc. The results reveal significant variability in model performance, with GPT-4 achieving a classification accuracy of 0.996 while LLaMA-3-8B lags behind at 0.758. These findings demonstrate the potential of LLMs in API classification, underscore the importance of careful model selection across different environments, and highlight the effectiveness of our system as an efficient and practical tool for customized API management.

\section{Related Work}
\subsection{LLMs for Natural Language Interfaces}
As machine learning field rapidly evolved over the past decade, its applications have expanded across diverse domains, revolutionizing industries such as technology  \cite{jin2024onlinelearningmultipletasks, tao2023meta}, healthcare \cite{gong2024graphicalstructurallearningrsfmri, bian2024diffusion, yang2024exploring, zheng2024identification, bian2022optimal, xintaoli}, finance \cite{zhao2024hedge} and road construction\cite{dan2024multiple}. These advancements not only address complex technical challenges but also offer streamlined solutions that improve user experiences \cite{zhu2023demonstration, songyukun, kang2022tie}. One of the most transformative breakthroughs has been the rise of natural language models. By harnessing deep learning techniques, natural language interfaces empower users to interact with complex systems through simple, intuitive commands\cite{zhang2024tfwttabularfeatureweighting, ji2023prediction, zeng2024wordepth, fan2024towards, zhang2024prototypical} , which has significantly lowered the barrier to traditionally technical operations, making them accessible to non-expert users. 

A prominent application is in data querying, where models can interpret natural language queries and convert them into structured commands like SQL, enabling non-technical users to retrieve information from databases without needing to understand the underlying syntax\cite{baig2022natural}. Beyond data querying, LLMs have been integrated into DevOps automation, allowing users to initiate and manage complex workflows using simple commands. This integration streamlines tasks such as infrastructure provisioning, deployment, and system monitoring, simplifying the traditionally complex domain of DevOps by providing a more user-friendly interaction model \cite{mehta2023automated}. 

Similarly, LLMs are being explored for their potentials in API interactions. While existing model like Codex can generate code snippets, the area of API call retrieval remains underdeveloped\cite{poesia2022synchromesh}. The complexity of APIs, which often involve diverse protocols, data formats, and domain-specific parameters, poses unique challenges. Therefore, there is a growing need to further research into the performance of LLMs in API retrieval scenarios. Such research would help identify the limitations of current models and refine their ability to accurately select and structure  API calls, ultimately making these tools more adaptable to real-world applications.

\subsection{Synthetis Data Generation}
To effectively evaluate LLMs' capabilities in these complex tasks, task-specific datasets tailored to the context of API retrieval are essential. Synthetic data generation has become a widely adopted approach in domains where real-world data is either scarce or sensitive. In the field of NLP, models like GPT-3 have been leveraged to quickly generate synthetic datasets for tasks such as text classification, enabling models to generalize to specialized domains more efficiently\cite{radford2019language}. Similarly, HumanEval leverages synthetic data to systematically evaluate models' abilities in code generation by providing tailored test cases \cite{chen2021evaluatinglargelanguagemodels}. In the healthcare domain, synthetic data is applied to simulate real-world conditions, allowing machine learning models to be trained and evaluated without compromising sensitive patient information \cite{chen2021evaluatinglargelanguagemodels}.
Despite progress in using synthetic data for general NLP tasks and other domains, API-specific dataset generation remains underexplored. Existing approaches often fail to provide domain-specific synthetic data tailored to API retrieval tasks, limiting their utility for benchmarking and evaluating LLM performance in real-world API interactions\cite{long2024llmsdrivensyntheticdatageneration}. Our work addresses this gap by developing a dataset generation pipeline that generates task-specific datasets tailored to API functions. This framework enables rapid benchmarking of various LLMs to assess their performance in API retrieval tasks. By automating the creation of synthetic datasets, our solution facilitates the quick and effective evaluation of LLMs, ensuring that models are tested in domain-relevant contexts.

\section{Methodology}
\begin{figure*}[ht]
  \centering
  \includegraphics[width=\linewidth]{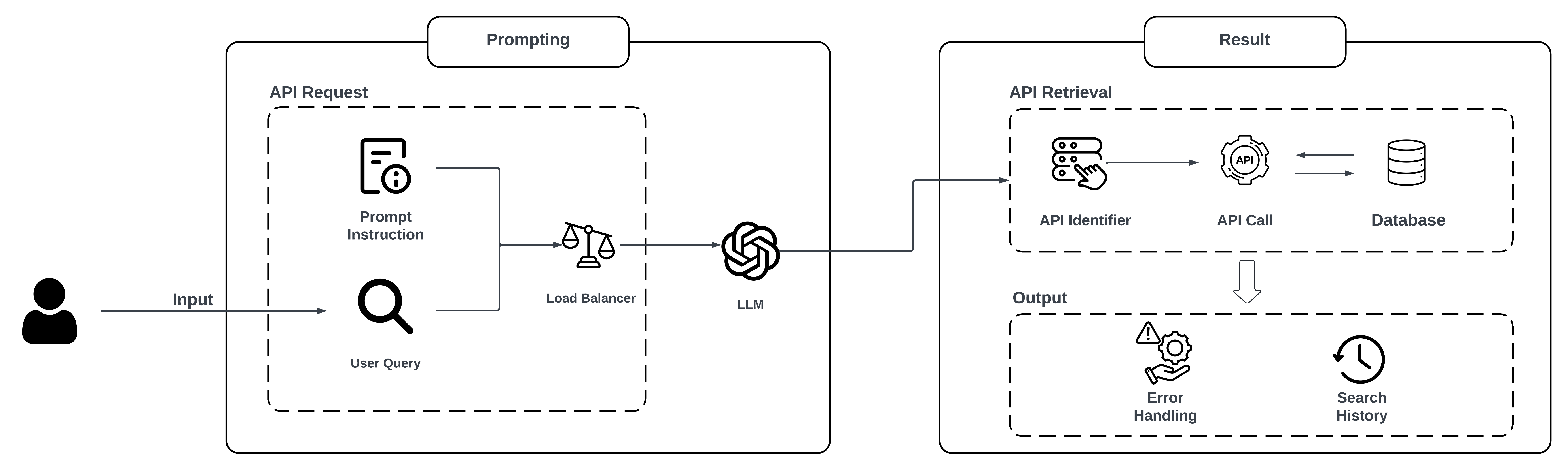}
  \caption{API Retrieval Framework}
  \label{fig:api_framework}
\end{figure*}
\begin{figure*}[ht]
  \centering
  \includegraphics[width=\linewidth]{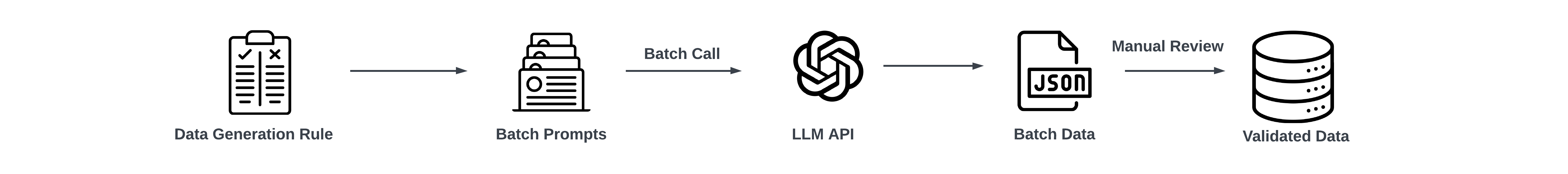}
  \caption{Dataset Generation}
  \label{fig:data_synthesis}
\end{figure*}
In our work, we propose an innovative framework to handle user prompts by identifying and utilizing the most suitable LLM for a specific application. Our methodology consists of two key components: an API retrieval system and a dataset generation pipeline. These components ensure efficient and accurate interaction with APIs, optimizing the selection of the best-performing LLM for the task.

\subsection{API Retrieval System}
API retrieval framework is an automated pipeline that effectively address user queries, ensuring that each query is correctly classified, passed to the appropriate API function, and the result is efficiently returned to the user. The structured workflow can be divided into the following key stages:

\subsubsection{\textbf{User Input}}
The system begins by receiving a natural language query from the user. During the prompting process, the user's input is combined with predefined prompt instructions before being fed into the LLM. These instructions define the API hierarchy and establish specific rules for the output format, ensuring the system's responses conform to the API's structural and functional requirements. This input can vary in complexity, ranging from simple questions to more complex commands. The flexibility of LLMs allows the system to interpret and handle a wide range of user inputs, even when the input is ambiguously phrased.

\subsubsection{\textbf{API Classification}}
Once the query is received, the integrated LLM maps it to appropriate API functions. Specifically, the LLM processes the query and returns a label that categorizes it based on the predefined API hierarchy. Relevant keywords required for API functions' input parameters are also retrieved. The label determines both the API modules and the specific function needed to fulfill the user’s request. To ensure efficient processing across available resources, a load balancer is applied to distribute incoming queries.

\subsubsection{\textbf{API Execution}}
After the label is returned, the API identifier will request servers for routes to render API functions. This step involves dynamically mapping the extracted keywords from the query to the input parameters required by API functions. For example, a weather API call might require the date and location, whereas a financial API might need specific transaction details. The API call is then executed, pulling data from the database. To maintain performance, Redis is used as a cache database to help alleviate load, ensuring smooth operations even under high I/O situations.

\subsubsection{\textbf{Result Returned}}
After the API call is processed, the result is returned in a user readable way. This step includes error handling. If any issues arise during API execution, such as invalid parameters or failed API calls, the system provides the user with relevant feedback. Additionally, a search history feature allows users to review their past queries, adding a layer of functionality for repeated interactions.

This end-to-end framework automates the entire API interactions, minimizing manual intervention and ensuring that user queries are handled efficiently and accurately. The integration of API hierarchy during prompting ensures that the system scales easily, allowing new API categories and functions to be added as needed.

\subsection{Dataset Generation Pipeline}

To efficiently evaluate and select the most suitable large language model (LLM) for API retrieval tasks, we proposes an automated data generation pipeline. The pipeline uses batch prompting to generate 100 unique synthetic queries per batch, simulating real-world user interactions with APIs. The batch calling method was chosen for its ability to optimize resource utilization by reducing the overhead associated with individual API requests, allowing multiple queries to be processed simultaneously. To further enhance the diversity of the queries and minimize repetition, we designed the prompts to instruct the LLM to vary phrasing and contexts for each query, while still adhering to the predefined API hierarchy. This ensures that the generated queries are rich in variability without sacrificing alignment with the intended API function. 

The generated queries are labeled with both the API modules and API functions, and stored in JSON format for efficient processing and management. To ensure high data quality and consistency throughout the process, each batch undergoes a manual review to verify label accuracy, ensuring the queries are correctly mapped to the appropriate API functions. The validated dataset is then used to evaluate LLMs, where each model is tasked with classifying the queries into the correct API modules and function. The model performance is measured using both module-level and function-level classification accuracy and the best performing model is integrated into the inference endpoint. By generating a high-quality, varied dataset to support comprehensive model evaluation, our approach ensures that the chosen LLM is both robust and adaptable, capable of handling diverse user queries in real-world API retrieval tasks.

\section{Experiments \& Results}
We conduct an experiment to assess the effectiveness of our system in generating high-quality datasets for API classification and in classifying API calls using various LLMs. The experiment is structured in two parts: dataset creation and API classification evaluation. 
\subsection{Dataset Generation}
\begin{figure*}[ht]
  \centering
  \includegraphics[width=\linewidth]{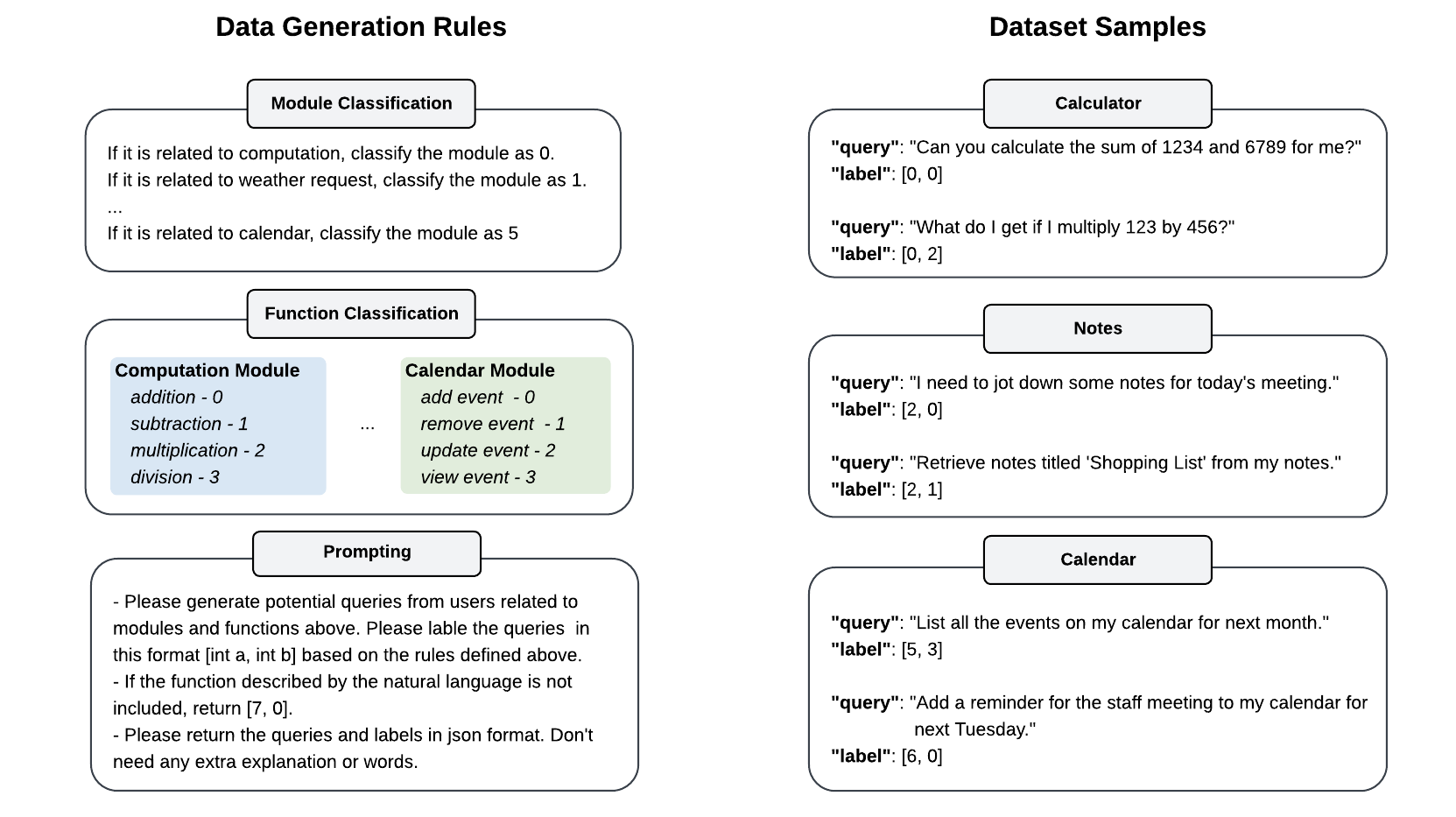}
  \caption{Data Generation Rules and Dataset Samples}
  \label{fig:dataset_illustration}
\end{figure*}
We generate datasets for six API modules commonly found in modern web and mobile applications: Calculator, Notes, Weather, Email, Notification, and Calendar. These modules are selected for their widespread applicability across various software systems, representing a diverse range of API functionalities. They span from basic arithmetic operations to more complex language-based tasks, ensuring comprehensive coverage of typical API interactions. For each of these modules, we simulate several commonly used functions, each corresponding to distinct operations or actions. For example, the calculator module includes functions such as \textit{add}, \textit{subtract}, \textit{multiply}, \textit{divide}, while the weather module encompasses functions like \textit{get\_today\_weather}, \textit{get\_weekly\_forecast}, and \textit{get\_air\_pollution}. By selecting API modules with multiple functions, we ensure that the dataset is sufficiently diverse to test the classification abilities of the LLMs across both simple and complex API operations. Additionally, we included a Routes-Not-Exist module to simulate invalid API calls, testing the system’s robustness in handling erroneous inputs. 

We employ GPT-4-turbo to generate the dataset for this experiment. GPT-4-turbo is an optimized version of GPT-4, known for its high accuracy in natural language processing and cost-effectiveness \cite{achiam2023gpt}. GPT-4-turbo is selected because of its ability to handle complex data generation tasks efficiently, allowing us to create a large-scale, high-quality dataset with minimal manual intervention. As outlined in the Methodology section, we pre-define specific Data Generation Rules to guide dataset generation for the selected API modules and functions. The Data Generation Rules serve as prompts to batch-call the GPT-4-turbo interface and generate labeled datasets in JSON format. The generated datasets are then manually reviewed with an accuracy of 99.9\%. The high accuracy demonstrates the reliability of leveraging LLMs to automatically generate high-quality datasets for complex API classification tasks, minimizing the need for extensive manual intervention. Figure \ref{fig:dataset_illustration} presents the Data Generation Rules we employ and sample entries from the generated datasets, where \textit{query} denotes the natural language prompt, the first entry of \textit{label} indicates the classified API module, and the second entry of \textit{label} specifies the corresponding function. The final dataset contains a total of 1300 samples. Table \ref{table:dataset_breakdown} provides a breakdown of the dataset, including simulated functions and the number of samples generated for each API module.

\renewcommand{\arraystretch}{1.3}
\begin{table*}[!htb]
\centering
\caption{LLM-Generated Sample Dataset Breakdown}
\label{table:dataset_breakdown}
\begin{tabular}{c|l|c}
\hline
\textbf{API Module} & \textbf{Functions} & \textbf{Total Samples} \\
\hline
Calculator  & \begin{tabular}[c]{@{}l@{}}\textit{add}, \textit{subtract}, \textit{multiply}, \textit{divide}, \textit{power}, \textit{log}, \textit{factorial}, \\
\textit{sin}, \textit{cos}, \textit{tan} \end{tabular} & 250                 \\
\hline
Weather & \begin{tabular}[c]{@{}l@{}} \textit{get\_today\_weather}, \textit{get\_weekly\_forecast}, \textit{get\_air\_pollution} \end{tabular}   & 200                   \\
\hline
Notes   & \begin{tabular}[c]{@{}l@{}} \textit{create}, \textit{get\_all\_notes}, \textit{delete\_note}, \textit{update\_note} \end{tabular}            & 200                   \\

\hline
Notification               & \begin{tabular}[c]{@{}l@{}}\textit{send\_notification}, \textit{view\_notification}, \textit{mark\_as\_read}, \\
\textit{delete\_notification} \end{tabular} & 200                   \\
\hline
Email        & \begin{tabular}[c]{@{}l@{}}\textit{compose\_email}, \textit{send\_email}, \textit{read\_email}, \\
\textit{reply\_email}, \textit{delete\_email}  \end{tabular} & 200                  \\
\hline
Calendar     & \begin{tabular}[c]{@{}l@{}} \textit{add\_event}, \textit{remove\_event}, \textit{update\_event}, \textit{view\_event}\end{tabular}   & 200                  \\
\hline
Routes Not Exist    &\textit{return\_invalid\_error}  & 50                   \\
\hline
\end{tabular}
\end{table*}

\subsection{API Classification Evaluation}
In this experiment, we evaluate a range of prominent LLMs, GPT-4, GPT-4o-mini, GPT-3.5-turbo, LLaMA3-70B, LLaMA3-8B, and Gemini-1.5, on the generated dataset to assess the effectiveness of different LLMs in processing complex natural language inputs and mapping them accurately to predefined API operations. These models are selected based on their accessibility, popularity, differing architectures and sizes, and varying performance capabilities, which allows us to compare their strengths and weaknesses in handling complex API classification tasks.

As described in the Methodology section, each selected LLM is integrated into the inference endpoint, where it processes the natural language queries from the generated dataset and classifies them into the appropriate API modules and functions. The inference results are then compared against the labeled data to measure model performance using two metrics: Module Level Classification Accuracy (MLC-Acc) and Function Level Classification Accuracy (FLC-Acc). MLC-Acc measures the model’s ability to correctly classify a query into the appropriate API module, such as identifying whether the query is related to the calculator, weather, email, or another module. MLC-Acc calculation is shown in Equation (\ref{eq:MLC_ACC}) below,
\renewcommand{\arraystretch}{1.3}
\begin{table*}[!ht]
\centering
\caption{Model Performance Across API Modules and Functions}
\label{table:acc_performance}
\begin{tabular}{c|cccccccc|c}
\hline
\multirow{2}{*}{\textbf{Model}} & \multicolumn{8}{c|}{\textbf{MLC-Acc $\uparrow$}} & \multicolumn{1}{c}{\textbf{FLC-Acc $\uparrow$}} \\
\cline{2-10}
                       & Calculator & Weather & Notes & Notification & Email & Calendar & Routes Not Exist & \textbf{Overall} & \textbf{Avg} \\
\hline
GPT-4                  & \textbf{0.996}      & \textbf{0.995}   & \textbf{0.985} & \textbf{0.990}        & \textbf{0.990} & \textbf{0.985}    & 1.000            & \textbf{0.992}   & \textbf{0.996}   \\
GPT-4o-mini            & 0.928      & 0.845   & 0.865 & 0.835        & 0.805 & 0.850    & 0.860            & 0.854   & 0.980   \\
GPT-3.5-turbo          & 0.976      & 0.930   & 0.915 & 0.910        & 0.895 & 0.920    & 0.900            & 0.925   & 0.984   \\
LLaMA3-70B             & 0.984      & 0.955   & 0.960 & 0.955        & 0.965 & 0.950    & 1.000            & 0.964   & 0.990   \\
LLaMA3-8B              & 0.868      & 0.775   & 0.740 & 0.715        & 0.690 & 0.725    & 0.800            & 0.758   & 0.948   \\
Gemini-1.5             & 0.980      & 0.965   & 0.960 & 0.975        & 0.925 & 0.925    & 1.000            & 0.958   & 0.981   \\
\hline
\end{tabular}
\end{table*}

\begin{equation}
    \label{eq:MLC_ACC}
    \text{MLC-Acc} = \frac{1}{N_m} \sum_{i=1}^{N_m} \mathbf{1} (\hat{y}_i^m = y_i^m)
\end{equation}

where $N_m$ is the total number of queries for a module, $y_i^m$ is the true module classification, $\hat{y}_i^m$ is the predicted module classification, and $\mathbf{1}$ is an indicator function that returns 1 if the predicted module matches the true module and 0 otherwise.

FLC-Acc measures how well the model can correctly classify the function within the correct API module, focusing on precision at the function level when the module classification is accurate. FLC-ACC is computed as shown in Equation (\ref{eq:FLC_ACC}),

\begin{equation}
    \label{eq:FLC_ACC}
    \text{FLC-Acc} = \frac{\sum_{i=1}^{N_f} \mathbf{1}(\hat{y}_i^m = y_i^m) \cdot \mathbf{1}(\hat{y}_i^f = y_i^f)}{\sum_{i=1}^{N_f} \mathbf{1}(\hat{y}_i^m = y_i^m)}
\end{equation}

where $N_f$ is the total number of function classifications, $y_i^f$ is the true function classification, $\hat{y}_i^f$ is the predicted function classification, $\sum_{i=1}^{N_f} \mathbf{1}(\hat{y}_i^m = y_i^m)$ counts how many times the module is correctly classified, and $\sum_{i=1}^{N_f} \mathbf{1}(\hat{y}_i^m = y_i^m) \cdot \mathbf{1}(\hat{y}_i^f = y_i^f)$ counts the number of cases where both the module and function are correctly classified.

\subsection{Results and Discussion}
Table \ref{table:acc_performance} summarizes the performance of each LLM across different API modules and function classifications. The results indicate that GPT-4 achieves the highest performance across both metrics, with an overall MLC-Acc of 0.992 and an average FLC-Acc of 0.996. It consistently achieves near-perfect accuracy in all modules, demonstrating the exceptional ability of GPT-4 to handle both module-level and function-level classification, making it highly reliable for real-world applications requiring high precision across various API modules. LLaMA3-70B follows closely, achieving an overall MLC-Acc of 0.964 and an average FLC-Acc of 0.990. LLaMA3-70B performs particularly well in modules like Calculator, Email, and Notification, indicating its robustness in both simpler and more complex tasks. Gemini-1.5 also demonstrates strong performance, with an overall MLC-Acc of 0.957 and an average FLC-Acc of 0.981. While Gemini-1.5 excels in most modules, its performance in modules like Email and Notification indicates that it may struggle with more nuanced or complex queries. However, it remains a reliable option for many API classification tasks, particularly when considering its strong performance in simpler tasks. GPT-3.5-turbo achieves moderate results, in both MLC-Acc and FLC-Acc. Smaller models, GPT-4o-mini and LLaMA3-8B, exhibit the most significant drop in accuracy, especially in complex modules. The large performance gap between LLaMA3-8B and its larger counterpart, LLaMA3-70B, suggests that model size directly impacts the model's capacity to understand complex language patterns, make accurate predictions, and generalize well across different contexts. While smaller models may be more computationally efficient, they often sacrifice accuracy, particularly in complex classification tasks. For API management systems requiring high precision, larger models like LLaMA3-70B and GPT-4 are far better suited.

The results underscore the strong capability of LLMs, particularly GPT-4 and LLaMA3-70B, in accurately classifying APIs across various common modules and functions, highlighting their potential to enhance API management systems. The experiment also demonstrates that our dataset generation framework provides an effective means to evaluate the performance of different LLMs, tailored to specific API functionalities and applications. By automating the generation and assessment process, our system enables comprehensive and scalable evaluation, making it easier to identify the most appropriate LLM for diverse API use cases. This approach not only streamlines the model selection but also ensures that applications are matched with the optimal model for their specific needs.

\section{Conclusion}
In this paper, we develop a comprehensive system that integrates LLMs for automating API classification and tailored dataset generation. The strong performance of GPT-4 and LLaMa-3-70B in our experiments indicates that LLMs can effectively understand and translate natural language inputs into precise API calls and improve both user interaction and API management. Our dataset generation approach provides a scalable solution for systematically assessing LLM capabilities across different API modules and functions, enabling developers or business owners to choose the most suitable model for specific tasks. Overall, the proposed system offers an efficient and adaptable method for integrating LLMs into API management for diverse applications.

Despite the promising results, our work still has some limitations. While larger LLMs like GPT-4 and LLaMA-3-70B offer high accuracy, their computational cost makes them less feasible for real-time applications in resource-constrained environments. Future research could explore finetuning and optimizing smaller LLMs for API classification without significantly sacrificing accuracy, making them more practical for a wider range of applications. Additionally, the current approach relies on predefined dataset generation rules, which may limit flexibility when adapting to highly dynamic or evolving API environments. Future work could focus on extending the framework to support dynamic dataset generation, enabling it to automatically adapt to new or evolving APIs without manual intervention.


\begin{thebibliography}{10}

\bibitem{vaswani2017attention}
Ashish Vaswani, Noam Shazeer, Niki Parmar, Jakob Uszkoreit, Llion Jones, Aidan~N. Gomez, \L{}ukasz Kaiser, and Illia Polosukhin.
\newblock Attention is all you need.
\newblock In {\em Proceedings of the 31st International Conference on Neural Information Processing Systems}, NIPS'17, page 6000–6010, Red Hook, NY, USA, 2017. Curran Associates Inc.

\bibitem{devlin2018bert}
Jacob Devlin.
\newblock Bert: Pre-training of deep bidirectional transformers for language understanding.
\newblock {\em arXiv preprint arXiv:1810.04805}, 2018.

\bibitem{zhang2022can}
Renrui Zhang, Ziyao Zeng, Ziyu Guo, and Yafeng Li.
\newblock Can language understand depth?
\newblock In {\em Proceedings of the 30th ACM International Conference on Multimedia}, pages 6868--6874, 2022.

\bibitem{wu2023bloomberggpt}
Shijie Wu, Ozan Irsoy, Steven Lu, Vadim Dabravolski, Mark Dredze, Sebastian Gehrmann, Prabhanjan Kambadur, David Rosenberg, and Gideon Mann.
\newblock Bloomberggpt: A large language model for finance.
\newblock {\em arXiv preprint arXiv:2303.17564}, 2023.

\bibitem{deng2024composerx}
Qixin Deng, Qikai Yang, Ruibin Yuan, Yipeng Huang, Yi~Wang, Xubo Liu, Zeyue Tian, Jiahao Pan, Ge~Zhang, Hanfeng Lin, et~al.
\newblock Composerx: Multi-agent symbolic music composition with llms.
\newblock {\em arXiv preprint arXiv:2404.18081}, 2024.

\bibitem{yuan2024rhyme}
Yixiao Yuan, Yangchen Huang, Yu~Ma, Xinjin Li, Zhenglin Li, Yiming Shi, and Huapeng Zhou.
\newblock Rhyme-aware chinese lyric generator based on gpt.
\newblock {\em arXiv preprint arXiv:2408.10130}, 2024.

\bibitem{liu2023text}
Shengchao Liu, Yanjing Li, Zhuoxinran Li, Anthony Gitter, Yutao Zhu, Jiarui Lu, Zhao Xu, Weili Nie, Arvind Ramanathan, Chaowei Xiao, et~al.
\newblock A text-guided protein design framework.
\newblock {\em arXiv preprint arXiv:2302.04611}, 2023.

\bibitem{li2024intelligent}
Xinjin Li, Yuanzhe Yang, Yixiao Yuan, Haowei Ni, Yu~Ma, and Yangchen Huang.
\newblock Intelligent vehicle classification system based on deep learning and multi-sensor fusion.
\newblock {\em Preprints}, 2024.

\bibitem{weng2024big}
Yijie Weng and Jianhao Wu.
\newblock Big data and machine learning in defence.
\newblock {\em International Journal of Computer Science and Information Technology}, 16(2), 2024.

\bibitem{tao2019fact}
Yiyi Tao, Yiling Jia, Nan Wang, and Hongning Wang.
\newblock The fact: Taming latent factor models for explainability with factorization trees.
\newblock In {\em Proceedings of the 42nd international ACM SIGIR conference on research and development in information retrieval}, pages 295--304, 2019.

\bibitem{yukun2019deep}
Yukun Song.
\newblock Deep {Learning} {Applications} in the {Medical} {Image} {Recognition}.
\newblock {\em American Journal of Computer Science and Technology}, 2(2):22--26, July 2019.

\bibitem{ji2024rag}
Yuelyu Ji, Zhuochun Li, Rui Meng, Sonish Sivarajkumar, Yanshan Wang, Zeshui Yu, Hui Ji, Yushui Han, Hanyu Zeng, and Daqing He.
\newblock Rag-rlrc-laysum at biolaysumm: Integrating retrieval-augmented generation and readability control for layman summarization of biomedical texts.
\newblock {\em arXiv preprint arXiv:2405.13179}, 2024.

\bibitem{calo2023leveraging}
Tommaso Cal{\`o} and Luigi De~Russis.
\newblock Leveraging large language models for end-user website generation.
\newblock In {\em International Symposium on End User Development}, pages 52--61. Springer, 2023.

\bibitem{fielding2000architectural}
Roy~Thomas Fielding.
\newblock {\em Architectural styles and the design of network-based software architectures}.
\newblock University of California, Irvine, 2000.

\bibitem{pautasso2008restful}
Cesare Pautasso, Olaf Zimmermann, and Frank Leymann.
\newblock Restful web services vs." big"'web services: making the right architectural decision.
\newblock In {\em Proceedings of the 17th international conference on World Wide Web}, pages 805--814, 2008.

\bibitem{liang2024taskmatrix}
Yaobo Liang, Chenfei Wu, Ting Song, Wenshan Wu, Yan Xia, Yu~Liu, Yang Ou, Shuai Lu, Lei Ji, Shaoguang Mao, et~al.
\newblock Taskmatrix. ai: Completing tasks by connecting foundation models with millions of apis.
\newblock {\em Intelligent Computing}, 3:0063, 2024.

\bibitem{song2023restgpt}
Yifan Song, Weimin Xiong, Dawei Zhu, Wenhao Wu, Han Qian, Mingbo Song, Hailiang Huang, Cheng Li, Ke~Wang, Rong Yao, et~al.
\newblock Restgpt: Connecting large language models with real-world restful apis.
\newblock {\em arXiv preprint arXiv:2306.06624}, 2023.

\bibitem{jin2024onlinelearningmultipletasks}
Yixin Jin, Wenjing Zhou, Meiqi Wang, Meng Li, Xintao Li, Tianyu Hu, and Xingyuan Bu.
\newblock Online learning of multiple tasks and their relationships: Testing on spam email data and eeg signals recorded in construction fields.
\newblock {\em arXiv preprint arXiv:2406.18311}, 2024.

\bibitem{tao2023meta}
Yiyi Tao.
\newblock Meta learning enabled adversarial defense.
\newblock In {\em 2023 IEEE International Conference on Sensors, Electronics and Computer Engineering (ICSECE)}, pages 1326--1330. IEEE, 2023.

\bibitem{gong2024graphicalstructurallearningrsfmri}
Yiru Gong, Qimin Zhang, Huili Zheng, Zheyan Liu, and Shaohan Chen.
\newblock {Graphical Structural Learning of rs-fMRI data in Heavy Smokers}.
\newblock {\em arXiv preprint arXiv:2409.08395}, 2024.

\bibitem{bian2024diffusion}
Wanyu Bian, Albert Jang, Liping Zhang, Xiaonan Yang, Zachary Stewart, and Fang Liu.
\newblock Diffusion modeling with domain-conditioned prior guidance for accelerated mri and qmri reconstruction.
\newblock {\em IEEE Transactions on Medical Imaging}, 2024.

\bibitem{yang2024exploring}
Yumeng Yang, Ashley Gilliam, Ethan~B Ludmir, and Kirk Roberts.
\newblock Exploring the generalization of cancer clinical trial eligibility classifiers across diseases.
\newblock {\em arXiv preprint arXiv:2403.17135}, 2024.

\bibitem{zheng2024identification}
Huili Zheng, Qimin Zhang, Yiru Gong, Zheyan Liu, and Shaohan Chen.
\newblock Identification of prognostic biomarkers for stage iii non-small cell lung carcinoma in female nonsmokers using machine learning.
\newblock {\em arXiv preprint arXiv:2408.16068}, 2024.

\bibitem{bian2022optimal}
Wanyu Bian, Yunmei Chen, and Xiaojing Ye.
\newblock An optimal control framework for joint-channel parallel mri reconstruction without coil sensitivities.
\newblock {\em Magnetic Resonance Imaging}, 89:1--11, 2022.

\bibitem{xintaoli}
Xintao Li and Sibei Liu.
\newblock Predicting 30-day hospital readmission in medicare patients: Insights from an lstm deep learning model.
\newblock {\em medRxiv}, 2024.
\newblock doi:10.1101/2024.09.08.24313212.

\bibitem{zhao2024hedge}
Siqiao Zhao, Zhikang Dong, Zeyu Cao, and Raphael Douady.
\newblock Hedge fund portfolio construction using polymodel theory and itransformer.
\newblock {\em arXiv preprint arXiv:2408.03320}, 2024.

\bibitem{dan2024multiple}
Han-Cheng Dan, Peng Yan, Jiawei Tan, Yinchao Zhou, and Bingjie Lu.
\newblock Multiple distresses detection for asphalt pavement using improved you only look once algorithm based on convolutional neural network.
\newblock {\em International Journal of Pavement Engineering}, 25(1):2308169, 2024.

\bibitem{zhu2023demonstration}
Yunyi Zhu, Cedric Honnet, Yixiao Kang, Junyi Zhu, Angelina~J Zheng, Kyle Heinz, Grace Tang, Luca Musk, Michael Wessely, and Stefanie Mueller.
\newblock Demonstration of chromocloth: Re-programmable multi-color textures through flexible and portable light source.
\newblock In {\em Adjunct Proceedings of the 36th Annual ACM Symposium on User Interface Software and Technology}, pages 1--3, 2023.

\bibitem{songyukun}
Yukun Song, Parth Arora, Rajandeep Singh, Srikanth~T. Varadharajan, Malcolm Haynes, and Thad Starner.
\newblock Going {Blank} {Comfortably}: {Positioning} {Monocular} {Head}-{Worn} {Displays} {When} {They} are {Inactive}.
\newblock In {\em Proceedings of the 2023 {International} {Symposium} on {Wearable} {Computers}}, pages 114--118, Cancun, Quintana Roo Mexico, October 2023. ACM.

\bibitem{kang2022tie}
Yixiao Kang, Zhenglin Zhang, Meiqi Zhao, Xuanhui Yang, and Xubo Yang.
\newblock Tie memories to e-souvenirs: Hybrid tangible ar souvenirs in the museum.
\newblock In {\em Adjunct Proceedings of the 35th Annual ACM Symposium on User Interface Software and Technology}, pages 1--3, 2022.

\bibitem{zhang2024tfwttabularfeatureweighting}
Xinhao Zhang, Zaitian Wang, Lu~Jiang, Wanfu Gao, Pengfei Wang, and Kunpeng Liu.
\newblock Tfwt: Tabular feature weighting with transformer.
\newblock {\em arXiv preprint arXiv:2405.08403}, 2024.

\bibitem{ji2023prediction}
Yuelyu Ji, Yuhe Gao, Runxue Bao, Qi~Li, Disheng Liu, Yiming Sun, and Ye~Ye.
\newblock Prediction of covid-19 patients’ emergency room revisit using multi-source transfer learning.
\newblock In {\em 2023 IEEE 11th International Conference on Healthcare Informatics (ICHI)}, pages 138--144. IEEE, 2023.

\bibitem{zeng2024wordepth}
Ziyao Zeng, Daniel Wang, Fengyu Yang, Hyoungseob Park, Stefano Soatto, Dong Lao, and Alex Wong.
\newblock Wordepth: Variational language prior for monocular depth estimation.
\newblock In {\em Proceedings of the IEEE/CVF Conference on Computer Vision and Pattern Recognition}, pages 9708--9719, 2024.

\bibitem{fan2024towards}
Xiaojing Fan and Chunliang Tao.
\newblock Towards resilient and efficient llms: A comparative study of efficiency, performance, and adversarial robustness.
\newblock {\em arXiv preprint arXiv:2408.04585}, 2024.

\bibitem{zhang2024prototypical}
Jinghan Zhang, Xiting Wang, Yiqiao Jin, Changyu Chen, Xinhao Zhang, and Kunpeng Liu.
\newblock Prototypical reward network for data-efficient rlhf.
\newblock {\em arXiv preprint arXiv:2406.06606}, 2024.

\bibitem{baig2022natural}
Muhammad~Shahzaib Baig, Azhar Imran, Aman~Ullah Yasin, Abdul~Haleem Butt, and Muhammad~Imran Khan.
\newblock Natural language to sql queries: A review.
\newblock {\em International Journal of Innovations in Science Technology}, 4:147--162, 2022.

\bibitem{mehta2023automated}
Deep Mehta, Kartik Rawool, Subodh Gujar, and Bowen Xu.
\newblock Automated devops pipeline generation for code repositories using large language models.
\newblock {\em arXiv preprint arXiv:2312.13225}, 2023.

\bibitem{poesia2022synchromesh}
Gabriel Poesia, Oleksandr Polozov, Vu~Le, Ashish Tiwari, Gustavo Soares, Christopher Meek, and Sumit Gulwani.
\newblock Synchromesh: Reliable code generation from pre-trained language models.
\newblock {\em arXiv preprint arXiv:2201.11227}, 2022.

\bibitem{radford2019language}
Alec Radford, Jeffrey Wu, Rewon Child, David Luan, Dario Amodei, Ilya Sutskever, et~al.
\newblock Language models are unsupervised multitask learners.
\newblock {\em OpenAI blog}, 1(8):9, 2019.

\bibitem{chen2021evaluatinglargelanguagemodels}
Mark Chen, Jerry Tworek, Heewoo Jun, Qiming Yuan, Henrique Ponde De~Oliveira Pinto, Jared Kaplan, Harri Edwards, Yuri Burda, Nicholas Joseph, Greg Brockman, et~al.
\newblock Evaluating large language models trained on code.
\newblock {\em arXiv preprint arXiv:2107.03374}, 2021.

\bibitem{long2024llmsdrivensyntheticdatageneration}
Lin Long, Rui Wang, Ruixuan Xiao, Junbo Zhao, Xiao Ding, Gang Chen, and Haobo Wang.
\newblock On llms-driven synthetic data generation, curation, and evaluation: A survey.
\newblock {\em arXiv preprint arXiv:2406.15126}, 2024.

\bibitem{achiam2023gpt}
Josh Achiam, Steven Adler, Sandhini Agarwal, Lama Ahmad, Ilge Akkaya, Florencia~Leoni Aleman, Diogo Almeida, Janko Altenschmidt, Sam Altman, Shyamal Anadkat, et~al.
\newblock Gpt-4 technical report.
\newblock {\em arXiv preprint arXiv:2303.08774}, 2023.

\end{thebibliography}
\end{document}